\documentclass{article}

  \PassOptionsToPackage{numbers, compress}{natbib}
\usepackage[preprint]{neurips_2020}
\usepackage[utf8]{inputenc} % allow utf-8 input
\usepackage[T1]{fontenc}  % use 8-bit T1 fonts
\usepackage{hyperref}    % hyperlinks
\usepackage{url}      % simple URL typesetting
\usepackage{booktabs}    % professional-quality tables
\usepackage{amsfonts}    % blackboard math symbols
\usepackage{nicefrac}    % compact symbols for 1/2, etc.
\usepackage{microtype}   % microtypography

\usepackage{bm}
\usepackage{times}
\usepackage{latexsym}
\usepackage{amsmath,amssymb,bbm,epsfig,bm}
\usepackage{algpseudocode}
\usepackage{balance}
\usepackage{multirow}
\usepackage{array}
\usepackage{tikz}
\usepackage{makecell}
\usepackage{graphicx}
\usepackage{caption}
\title{CAPT: Contrastive Pre-Training for Learning Denoised Sequence Representations}

\author{
 Fuli Luo$^{*}$ \\
 Peking University \\
 \texttt{luofuli@pku.edu.cn} \\
  \And
  Pengcheng Yang$^{*}$ \\
  Peking University \\
  \texttt{yang\_pc@pku.edu.cn} \\
  \And
  Shicheng Li \\
  Peking University \\
  \texttt{lisc99@pku.edu.cn} \\
  \And
  Xuancheng Ren \\
  Peking University \\
  \texttt{renxc@pku.edu.cn} \\
  \And
  Xu Sun \\
  Peking University \\
  \texttt{xusun@pku.edu.cn} \\
}

\begin{document}

\maketitle
\renewcommand{\thefootnote}{\fnsymbol{footnote}}
\footnotetext[1]{Equal contribution.}
\renewcommand{\thefootnote}{\arabic{footnote}}

\begin{abstract}
Pre-trained self-supervised models such as BERT have achieved striking success in learning sequence representations, especially for natural language processing. These models typically corrupt the given sequences with certain types of noise, such as masking, shuffling, or substitution, and then try to recover the original input. However, such pre-training approaches are prone to learning representations that are covariant with the noise, leading to the discrepancy between the pre-training and fine-tuning stage. To remedy this, we present ContrAstive Pre-Training (CAPT) to learn noise invariant sequence representations. The proposed CAPT encourages the consistency between representations of the original sequence and its corrupted version via unsupervised instance-wise training signals. In this way, it not only alleviates the pretrain-finetune discrepancy induced by the noise of pre-training, but also aids the pre-trained model in better capturing global semantics of the input via more effective sentence-level supervision. Different from most prior work that focuses on a particular modality, comprehensive empirical evidence on 11 natural language understanding and cross-modal tasks illustrates that CAPT is applicable for both language and vision-language tasks, and obtains surprisingly consistent improvement, including 0.6\% absolute gain on GLUE benchmarks and 0.8\% absolute increment on $\text{NLVR}^2$.

\end{abstract}

\section{Introduction}

Recently, pre-trained self-supervised models such as BERT~\cite{devlin2019bert} have attracted an increasing amount of attention in natural language processing and vision-language processing. 
Benefiting from common knowledge contained in massive unlabeled data~\cite{liu2019roberta}, the pretraining-finetuning framework has become a representative paradigm for advancing various language-related downstream tasks. 

Most endeavors on pre-trained representation models rely on elaborately designed self-supervised tasks, which typically corrupt the given sequence with certain types of noise (e.g., masking in BERT~\cite{devlin2019bert} or shuffling in BART~\cite{lewis2019bart}), and then train the model to recover the original sequence. 
As a consequence, the learned representations tend to be covariant with the input noise of pre-training in this paradigm. 
However, when transferred to downstream tasks, the pre-trained model is responsible for encoding the original sequence without noise, and is expected to obtain noise invariant representations. 
Such pretrain-finetune discrepancy not only impedes fast fine-tuning, but also may result in suboptimal sequence representations, thus affecting the performance in downstream tasks.

%%%%%%%%%%%%
% \begin{table}[t]
%   \caption{Noise types used in the current natural language (upper) and vision-language (lower) sequence representation models.}
% 	\label{tab:noisetype}
% 	\vskip 0.15in
% 	\centering
% 	\small
% 	\begin{tabular}{l l}
% 		\toprule
% 		\textbf{Models}  & \textbf{Noise type}       \\ \midrule
% 		BERT~\cite{devlin2019bert}           & Mask tokens       \\
% 		SpanBERT~\cite{joshi2019spanbert}      & Mask spans      \\
% % 		RoBERTa~\cite{liu2019roberta}         & Mask token       \\
% % 		XLNet~\cite{yang2019xlnet}          & Shuffle token     \\
% 		ELECTRA~\cite{clark2020electra}       & Replace tokens       \\
% 		StructBERT~\cite{wang2019structbert}     & Mask + Shuffle tokens   \\ 
% % 		BART~\cite{lewis2019bart} & Mask + Shuffle + Replace. \\
% 		\midrule
% 		UNITER~\cite{chen2019uniter}       & Mask tokens + regions    \\
% 		LXMERT~\cite{tan2019lxmert}       & Mask tokens + regions     \\ 
% 		\bottomrule
% 	\end{tabular}
% 	\vskip -0.1in
% \end{table}
%%%%%%%%%%%%

%%%%%%%%%%%%
% \begin{figure}[ht]
% % \vskip 0.1in
% \centering
% \includegraphics[width=0.36\textwidth]{model-overview.pdf}
% % \caption{The illustration of the ContrAstive Pre-Training (CAPT).
% % }
% \caption{ContrAstive Pre-Training (CAPT) encourages the representations of inputs sharing semantics like ($\bm x$, $\bm{\hat{x}}$) to be similar, while penalizing the representations of inputs expressing different semantics like ($\bm{x}$, $\bm{y}$) and ($\bm{\hat{x}}$, $\bm y$) to be distant.
% }
% \label{fig:method-overview}
% \vskip -0.2in
% \end{figure}
%%%%%%%%%%%%

To remedy this, we present ContrAstive Pre-Training (CAPT) to learn noise invariant (or denoised) sequence representations.
%, inspired by the Noise Contrastive Estimation~\cite{Michael2010NCE}.
The core idea of CAPT is to enhance the consistency between semantic representations of the original sequence and that of corresponding corrupted version (e.g. the masked sequence) via unsupervised instance-wise training signals.
%can be fully utilized via elaborately designed semantic contrastive loss.
%As shown in Figure~\ref{fig:method-overview}, our approach 
In more detail, it strives to pull the representation of the corrupted sequence towards that of the original instance in the semantic space, while pushing it away from representations of other instances.
% Such training objectives are formulated as a multi-class classification task, which aims at classifying the original sequence to the class of its corrupted version and vice versa, while classifying different instances into different classes.
% For implementation feasibility, two effective model extension are proposed to further enhance the capability of the model to extract noise-concentrated and instance-diffused features.
Moreover, in order to enable the model to learn from more ``difficult'' and ``diverse'' instances, two effective methods are proposed to further enhance the capability of the model to extract noise-concentrated and instance-diffused features.
With such training objective, the pre-trained model is encouraged to learn noise invariant representations, thereby alleviating the pretrain-finetune discrepancy to some extent.

As an additional benefit, CAPT also assists the pre-trained model to more effectively capture the global semantics of the input. 
Most prior work only focuses on token-level pre-training tasks (e.g. masked language modeling), which lacks the modeling of global semantics of the input. 
Some other efforts alleviate this problem by introducing sentence-level pre-training tasks (e.g. next sentence prediction) that rely on the relative position of segments in the document.
However, the semantic connection between these segments tends to be excessively loose, which may result in confusing gradient signals~\cite{liu2019roberta}. 
By contrast, our CAPT offers incentives for representations of inputs sharing the same semantics (the original instance and its corrupted version) to be similar, while the representations of inputs expressing different semantics (different instances) are penalized to be distinguished from each other.
Such more reasonable sentence-level supervision enables our approach to look beyond the local structures of input sequences and become more aware of the global semantics.
%With such more reasonable sentence-level supervision, our approach achieves better modeling of global semantics of the input.

We perform the evaluation on a comprehensive suite of benchmark, covering 8 natural language understanding and 3 cross-modal tasks. 
Extensive empirical evidence demonstrates that our approach can achieve consistent improvements over the baselines in both language and vision-language domains.
To be more specific, our CAPT raises the performance of RoBERTa~\cite{liu2019roberta} from 88.9\% to 89.5\% on the GLUE dev set, and also surpasses LXMERT~\cite{tan2019lxmert} by 0.5\%, 0.6\% and 0.8\% on VQA, GQA and $\text{NLVR}^2$, respectively.

\section{Methodology}

\subsection{Contrastive pre-training}
\label{sec:capt_overview}

As a general framework, our CAPT can be built on various pre-trained models in either language or vision-language domains.
Therefore, we use the symbol $\mathcal{E}$ to represent a series of generalized pre-trained models. 
Different from prior work that tends to learn representations covariant with the noise of pre-training, CAPT aims at aiding $\mathcal{E}$ in learning noise invariant sequence representations by enhancing the consistency between representations of the original sequence and its corrupted version.

Specifically, for pre-trained model $\mathcal{E}$ and any sequence $\bm x$, the model-specific noise (e.g. masking in BERT) can be added to $\bm x$ to construct its corrupted version $\bm{\hat{x}}$.
Then, the pre-trained model $\mathcal{E}$ encodes $\bm x$ or $\bm{\hat{x}}$ with self-attention mechanism~\cite{vaswani2017attention} to obtain hidden representations $\bm{h}(\bm x)=\mathcal{E}(\bm x)$ or $\bm{h}(\bm{\hat{x}})=\mathcal{E}(\bm{\hat{x}})$. Both $\bm{h}(\bm x)$ and $\bm{h}(\bm{\hat{x}})$ belong to the representation space $\mathbb{R}^{m \times d}$, where $m$ denotes the length of the input sequence and $d$ is the dimension of hidden representation.

\begin{figure*}[t]
\vskip -0.15in
\centering
\includegraphics[width=1.0\textwidth]{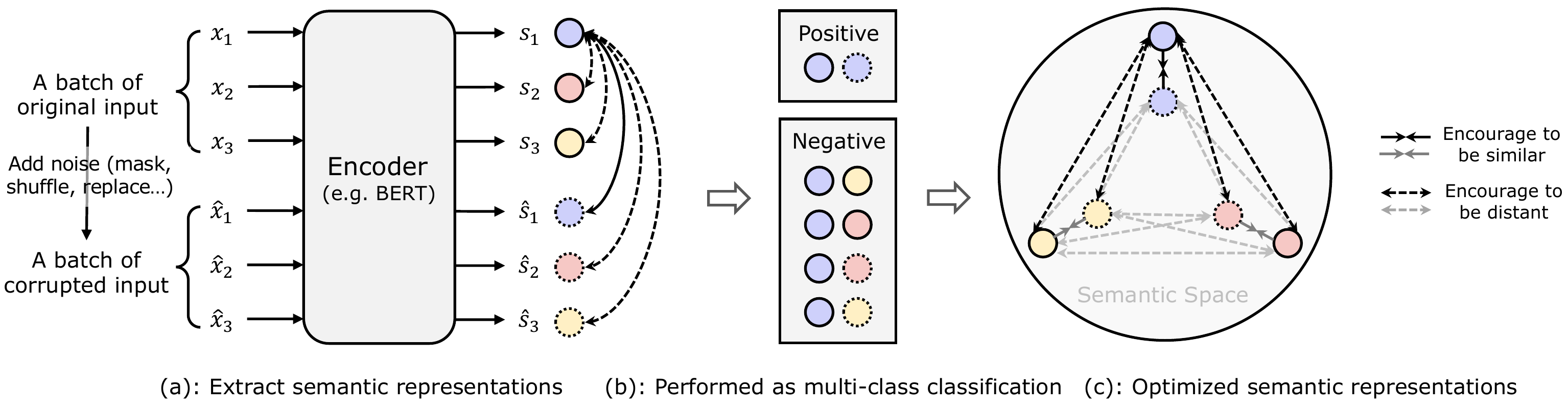}
\vskip -0.05in
\caption{ContrAstive Pre-Training (CAPT) performed in a training batch. 
(a): A batch of original sequences and corresponding corrupted versions are fed into the encoder to extract global representations. (b, c): CAPT aims at classifying the original sequence (e.g. $\bm x_1$) to the class of its corrupted version (e.g. $\bm{\hat{x}_1}$) and vice versa to encourage them to be similar in the semantic space, while classifying different instances into other classes to encourage them to be distant (see Eq.~(\ref{eq:nce_x})).}
\label{fig:method}
\vskip -0.05in
\end{figure*}

Different from prior work~\cite{devlin2019bert}, we apply an extra aggregation layer $\mathcal{A}$ to obtain the global semantic representation of the input. 
Here $\mathcal{A}$ can be implemented as a multi-layer perceptron with the representation of special classification token or the mean-pooling of all token representations as input.
The final global semantic representations $\bm{s}(\bm{x}) \in \mathbb{R}^{d}$ and $\bm{s}(\bm{\hat{x}}) \in \mathbb{R}^{d}$ of $\bm x$ and $\bm{\hat{x}}$ are computed as:
\begin{equation}
\label{eq:hidden}
% \small
% \begin{split}
{\bm s}(\bm x) = \ell_2^{\rm norm} \circ \mathcal{A}\big(\bm{h}(\bm x)\big), \quad  
{\bm s}(\bm{\hat{x}}) = \ell_2^{\rm norm} \circ \mathcal{A}\big(\bm{h}(\bm{\hat{x}})\big)
\end{equation}
where $\ell_2^{\rm norm}(\cdot)$ represents $\ell_2$-normalization and $\circ$ denotes the composition of operations.
In order to obtain noise invariant sequence representations, we expect ${\bm s}(\bm x)$ and ${\bm s}(\bm{\hat{x}})$ to be as similar as possible, which can also be derived from the characteristic that $\bm x$ and $\bm{\hat{x}}$ share semantics. 
At the same time, the global semantic representations of different instances should be distinguished from each other to extract the high-level specific signals of the input.
Motivated by this, we employ contrastive loss~\cite{hadsell06contrastive}
to model such training objectives, which can be formalized as a multi-class classification task. 
We represent a training batch of the original sequences as $\{\bm x_1, \cdots, \bm x_n\}$ where $n$ is the batch size, and its corresponding corrupted data is denoted as $\{\bm{\hat{x}}_1, \cdots, \bm{\hat{x}}_n\}$.
% Then, the training loss for the original sequence $\bm{x}_i$ is defined as:
Intuitively, the loss should be low when $s_i$ is similar to its corrupted version $\hat{s}_i$ (positive example) and dissimilar to all other inputs (negative examples).
Thus, the training loss for the original sequence $\bm{x}_i$ is defined as:
\begin{equation}
  \label{eq:nce_x}
  \mathcal{L}(\bm{x}_i) = -{\rm log}\frac{{\rm exp}({\bm s}_i \cdot \bm{\hat{s}}_i / \tau)}{\sum\limits_{j}{\rm exp}({\bm s}_i \cdot \bm{\hat{s}}_j / \tau) + \sum\limits_{j\not=i}{\rm exp}({\bm s}_i \cdot {\bm s}_j / \tau)}
\end{equation}
where ${\bm s}_i={\bm s}(\bm x_i)$, $\bm{\hat{s}}_i={\bm s}(\bm{\hat{x}}_i)$, and $\tau$ is the temperature presented in Section~\ref{sec:model_extension}. Similarly, the training loss for the corrupted sequence $\bm{\hat{x}}_i$ can be defined as:
\begin{equation}
  \label{eq:nce_x_hat}
  \mathcal{L}(\bm{\hat{x}}_i) = -{\rm log}\frac{{\rm exp}(\bm{\hat{s}}_i \cdot {\bm s}_i / \tau)}{\sum\limits_{j}{\rm exp}(\bm{\hat{s}}_i \cdot {\bm s}_j / \tau) + \sum\limits_{j\not=i}{\rm exp}(\bm{\hat{s}}_i \cdot \bm{\hat{s}}_j / \tau)}
\end{equation}
Eq.~(\ref{eq:nce_x}) and Eq.~(\ref{eq:nce_x_hat}) essentially correspond to the log loss of a softmax-based classifier measuring semantic similarity by dot product. The classifier treats each instance as a distinct class, and aims to classify $\bm{x}_i$ to the class of $\bm{\hat{x}}_i$ and vice versa. 
More vividly, as shown in Figure~\ref{fig:method}, Eq.~(\ref{eq:nce_x}) and Eq.~(\ref{eq:nce_x_hat}) strive to pull the original representation $\bm s_i$ towards the representation $\bm{\hat{s}}_i$ of the corrupted sequence $\bm {\hat{x}}_i$, and push it away from global semantic representations of other sequences.
By maximizing the semantic similarity of global representations of $\bm x_i$ and $\bm{\hat{x}}_i$, the model is encouraged to learn noise-invariant and instance-diffused representations.
On this account, the self-supervised representation model is pre-trained in a manner that is more applicable for noise-free data distribution. 
This alleviates the pretrain-finetune discrepancy induced by the noise of pre-training to some extent, leading to improved performance on downstream scenarios. 
Besides, by introducing more reasonable sentence-level supervision, our approach can also capture global semantics of the input more effectively.

For the original training batch $\{\bm x_1, \cdots, \bm x_n\}$ and the constructed corrupted inputs $\{\bm{\hat{x}}_1, \cdots, \bm{\hat{x}}_n\}$, the final contrastive loss is the total sum of losses of all instances, which can be formulated as:
\begin{equation}
  \label{eq:capt_total}
  \mathcal{L}_{\rm capt} = \sum\nolimits_{i=1}^n \big\{\mathcal{L}(\bm x_i) + \mathcal{L}(\bm{\hat{x}}_i)\big\}
\end{equation}

\subsection{Model extensions}
\label{sec:model_extension}
% We improve the proposed CAPT methods by proposing two extensions to the model: adaptive temperature and memory queue.

\paragraph{Adaptive temperature.} 
Prior work \cite{simclr} has illustrated that the temperature $\tau$ controlling the concentration level of the sample distribution in Eq.~(\ref{eq:nce_x}) and Eq.~(\ref{eq:nce_x_hat}) exhibits significant impact on model performance.
%\textcolor[rgb]{1,0,0}{
A suitable temperature can help the model learn from hard negatives via the gradient \cite{simclr}.
Thus, it needs to be tuned elaborately to obtain a satisfactory fixed value. 
However, the optimal $\tau$ is constantly evolving as training proceeds. 
In fact, the gradient of the loss function Eq.~(\ref{eq:nce_x}) with respect to the representation $\bm{s}_i$ can be derived as:
\begin{equation}
  \label{eq:gradient}
  \frac{1}{\tau Z}
  \Bigg[
  \Big({\rm exp}(\bm{s}_i\cdot\bm{\hat{s}}_i/\tau)-Z\Big)\bm{\hat{s}}_i + 
  \sum\limits_{j\not=i}\Big(
  {\rm exp}(\bm{s}_i\cdot\bm{s}_j/\tau)\bm{s}_j + 
  {\rm exp}(\bm{s}_i\cdot\bm{\hat{s}}_j/\tau)\bm{\hat{s}}_j\Big)
  \Bigg]
\end{equation}
where $Z=\sum\nolimits_{j}{\rm exp}({\bm s}_i \cdot \bm{\hat{s}}_j / \tau) + \sum\nolimits_{j\not=i}{\rm exp}({\bm s}_i \cdot {\bm s}_j / \tau)$ is the normalization factor.
We found that the norm of the gradient in Eq.~(\ref{eq:gradient}) tends to be inversely proportional to $\tau$. 
% Ideally, the model update should be carefully controlled both in the early stage of contrastive pre-training to stabilize the training, and when approaching convergence to avoid getting stuck in local optima. 
Ideally, the model update should be carefully controlled not both in the early stage of contrastive pre-training to stabilize the training, but also in the later stage to avoid ending up bouncing around the minimum or getting stuck in local optima. 
Therefore, we implement the temperature $\tau$ as the following inverted triangle schedule instead of a predetermined fixed value:
\begin{equation}
\tau(t)=\frac{1}{{\rm T}}\left| t-\frac{{\rm T}}{2} \right|+0.05
\end{equation}
where $t$ denotes the learning step and $\rm T$ refers to the preset total number of updates.

\paragraph{Memory queue.}
As depicted in Figure~\ref{fig:method}, we treat different instances from the same batch as negative samples. Several recent studies~\cite{simclr,dai2017contrastive} have illustrated that contrastive learning can benefit from larger negative representations.
%batch sizes since it enlarge the number of negative examples.
% However, the GPU memory size decides the batch size, thus limiting the size of negative examples. 
% As depicted in \cite{simclr, dai2017contrastive, he2019momentum}, contrastive learning requires plenty of negative samples to capture the instance-diffused high-level features.
However, due to massive model parameters and limited machine memory, implementation with large batch size tends to be infeasible in many circumstances. 
To remedy this, we employ a dynamic memory queue $\mathcal{Q}$ to store the desired negative representations~\cite{he2019momentum}. At each learning step, the aggregated representations of the current batch of original inputs and its corresponding corrupted inputs are enqueued into $\mathcal{Q}$. Once reaching the preset capacity of $\mathcal{Q}$, the oldest redundant representations are dequeued. Different from~\cite{wu2018unsupervised}, the negative representations stored in $\mathcal{Q}$ are updated along with the training process to provide competitive confounders and informative signals for positive instances.

\section{Experiments on language tasks}
\label{sec:textual_experiments}
%This section presents experiments on language tasks, including specific implementation and detailed results.

\subsection{Implementation}
\label{sec:textual_implementation}
% %For learning language sequence representations, 

% For each input sequence, we mask 15\% 
% In our implementation, following BERT, 
% %for any sentence $\bm w$ in the training data, 
% we randomly select 15\% tokens of $\bm w$ as candidates for performing masking.

For learning language representations, the noise corrupting the input sentence $\bm x$ can be implemented as masking like BERT or shuffling like BART. 
% In our implementation, the main experiments follow the corrupting way like BERT.
In our implementation, the main experiments follow the same corruption as BERT.
That is, we randomly mask 15\% tokens of $\bm x$ to construct its corrupted version $\bm{\hat{x}}$. 
Then, both $\bm x$ and $\bm{\hat{x}}$ in the training batch are fed into the encoder to compute the CAPT loss and mask language model (MLM) loss simultaneously. The final training loss is the sum of the above two.
% And then we feed $(x, \hat{x})$ into the same batch to compute the CAPT loss and mask language model (MLM) loss simultaneously.
% The final training objective is the sum of them.
% \footnote{The reason is that the MLM excels at learning representations conditioned on masking noise and CAPT can help to learning a noise invariant representation.}
More analysis of the influence of other corruption approaches can be found in Section~\ref{sec:noise}.

During pre-training, we train a small model (Section~\ref{sec:ablation_study}) to validate the influence of key components of CAPT, and a large model (Section~\ref{sec:textual_results} and Section~\ref{sec:noise}) to demonstrate the effectiveness of CAPT for learning denoised text representation at a large scale. 

The small model, which is designed as a 6-layer Transformer with 256 hidden size and 4 attention heads, is trained on BookCorpus and English Wikipedia datasets. 
For the large CAPT model, we adopt RoBERT-Large model and training settings, which is a 24-layer Transformer with 1024 hidden size, 16 attention heads, and is trained on larger datasets.

% For a fair comparison, we adopt the same training corpora as ELECTRA-Small and RoBERT-Large for our two sized CAPT models, respectively. 
% For the small CAPT model, we follow a similar model size as~\cite{clark2020electra}, which is a 6-layer Transformer with 256 hidden size, 4 attention heads. For the large CAPT model, we adopt RoBERT-Large settings, which is a 24-layer Transformer with 1024 hidden size, 16 attention heads.
% For a fair comparison, we adopt the same training corpora as ELECTRA-Small and RoBERT-Large for our two sized CAPT models, respectively. 
Readers can refer to \cite{liu2019roberta} for the statistics of the dataset and processing details.
The aggregation layer $\mathcal{A}$ that takes the representation of special classification token as input is implemented as a nonlinear projection with one hidden layer. The inner hidden size of $\mathcal{A}$ is set to the same as FFN inner hidden size and the output hidden size is set to the same as Transformer hidden size.
The queue size is set to 8192 and we use Adam optimizer.
%and we use Adam optimizer.
%A byte-level BPE vocabulary which contains 50K sub-word units is used to preprocess the training corpora.
%We use Adam optimizer with $\beta_1$ = 0.9, $\beta_2$ = 0.98, and $\epsilon$ = 1e-6. 
The peak learning rate with linear warmup and decay is set to 5e-4 and 6e-4 for small and large models, respectively\footnote{More details about pre-training hyper-parameters can be found in Appendix A.}.

\subsection{Evaluation tasks}
\label{sec:textual_tasks}

%We perform the evaluation on extensive downstream tasks to demonstrate the effectiveness and versatility of our approach. The evaluation tasks are also divided into two broad categories: language understanding tasks including GLUE and SQuAD, and vision-language tasks including VQA, GQA, and $\text{NLVR}^2$.
%\paragraph{GLUE}
We perform the evaluation on the General Language Understanding Evaluation (GLUE) benchmark~\cite{wang2019glue}. Following previous work~\cite{devlin2019bert,liu2019roberta}, we experiment on 8 natural language understanding tasks, including linguistic acceptability (CoLA), sentiment analysis (SST), text paraphrase (MRPC and QQP), sentence similarity (STS-B), and natural language inference (MNLI, QNLI, and RTE). 

For fine-tuning, we adopt the same settings and hyper-parameters as those of RoBERTa. All GLUE tasks are framed as single-sentence or sentence-pair classification tasks, except for STS-B which is a regression task. 
Extra multi-layer perceptrons (MLP) are added to perform classification or regression with the representation of special classification token as the input. 
The evaluation metrics include Matthews correlation for CoLA, Pearson correlation for STS-B, and accuracy for other tasks. 

%\paragraph{SQuAD}
%The Stanford Question Answering Dataset (SQuAD) is a reading comprehension dataset based on articles in Wikipedia. Given a question and a context paragraph, the task aims to predict the answer span in the paragraph. We evaluate our model on two versions of SQuAD: v1.1~\cite{rajpurkar16squad1} and v2.0~\cite{rajpurkar2018squad2}. SQuAD v1.1 consists of 100,000 question-context pairs where the context always contains an answer, while SQuAD v2.0 introduces 50,000 unanswerable questions in addition. The predictions are evaluated in terms of exact match (EM) and F1 score.

%%%%%%%%%%%%%%%%%%%% table: glue %%%%%%%%%%%%%%%%%%%%%%%%%%%%%%
\begin{table*}[t]
\footnotesize
\renewcommand{\arraystretch}{1.2}
\centering
\vskip 0.15in
\begin{tabular*}{\textwidth}{l@{\extracolsep{\fill}}cccccccc|c}
\toprule
% \multirow{2}{*}{\bf Model} & \textbf{CoLA} & \textbf{SST-2} & \textbf{MRPC} & \textbf{STS-B} & \textbf{QQP} & \textbf{MNLI} & \textbf{QNLI} & \textbf{RTE} & \multirow{2}{*}{\bf Average } \\
% & Mcc & Acc & Acc & Pcc & Acc & Acc & Acc & Acc & \\
% & 8.5k/Mcc & 67k/Acc & 3.5k/Acc & 5.7k/Pcc & 363k/Acc & 392k/Acc & 108k/Acc & 2.5k/Acc & \\
{\bf Model} & \textbf{CoLA} & \textbf{SST-2} & \textbf{MRPC} & \textbf{STS-B} & \textbf{QQP} & \textbf{MNLI} & \textbf{QNLI} & \textbf{RTE} & {\bf Avg } \\

\midrule 
BERT~\cite{devlin2019bert} & 60.6 & 93.2 & 88.0 & 90.0 & 91.3 & 86.6 & 92.3 & 70.4 & 84.0 \\
XLNet~\cite{yang2019xlnet} & 63.6 & 95.6 & 89.2 & 91.8 & 91.8 & 89.8 & 93.9 & 83.8 & 87.4 \\
% \textbf{RoBERTa-100K} & 66.1 & 95.6 & 91.4 & 92.2 & 92.0 & 89.3 & 94.0 & 82.7 & 87.9 \\
RoBERTa~\cite{liu2019roberta} & 68.0 & 96.4 & 90.9 & 92.4 & 92.2 & 90.2 & 94.7 & 86.6 & 88.9 \\
% ELECTRA~\cite{clark2020electra} & \bf 69.3 & 96.0 & 90.6 & 92.1 & \textbf{92.4} & 90.5 & 94.5 & 86.8 & 89.0 \\
CAPT (Ours) & \textbf{69.2} & \textbf{96.5} & \textbf{92.1} & \textbf{92.5} & \textbf{92.3} & \textbf{90.7} & \textbf{95.0} & \textbf{88.0} & \underline{\textbf{89.5}} \\
\midrule
\multicolumn{10}{l}{\textit{Test set results for single models (no ensembling)}} \\
BERT~\cite{devlin2019bert} & 60.5 & 94.9 & 85.4 & 87.6 & 89.3 & 86.7 & 92.7 & 70.1 & 83.4 \\
SpanBERT~\cite{joshi2019spanbert} & \textbf{64.3} & 94.8 & 87.9 & 89.9 & 89.5 & 88.1 & 94.3 & 79.0 & 86.0 \\
% RoBERTa & 63.8 & 96.3 & 88.1 & \textbf{91.9} & \textbf{90.0} & 89.8 & \textbf{94.8} & 86.0 & 87.6 \\
CAPT (Ours) & 64.1 & \textbf{96.8} & \textbf{88.9} & \textbf{91.5} & \textbf{89.7} & \textbf{90.0} & \textbf{94.7} & \textbf{86.9} & \underline{\textbf{87.8}} \\
\bottomrule
\end{tabular*}
\caption{GLUE dev and test results of large models (24-layer transformer). We only list the dev and test results on each set that are available in the published papers.
% The test results are from the public GLUE leaderboard, except that RoBERTa test results are our resubmission of a single model.
``Avg'' denotes the average score in terms of the reported metrics, which is slightly different from that in the GLUE leaderboard.
}
\label{tab:glue}
\vskip -0.05in
\end{table*}
%%%%%%%%%%%%%%%%%%%% table: glue %%%%%%%%%%%%%%%%%%%%%%%%%%%%%%

\subsection{Results}
\label{sec:textual_results}
Following prior work~\cite{liu2019roberta,yang2019xlnet}, we report results on both dev and test data sets. 
For the dev set, we report the median of multiple random fine-tuning runs like RoBERTa to show reliable results. 
For the test set, since the ground-truth labels are not obtainable, we only made a single-model submission to the GLUE evaluation server. 
Note that most systems on the GLUE leaderboard adopt different ensemble metrics and task-specific fine-tuning methods (e.g. formulating QNLI as a ranking task or using multi-task fine-tuning), increasing the difficulty for a direct comparison. Thus, following \cite{devlin2019bert}, we only report non-ensemble single-task results.
%\footnote{\url{https://github.com/pytorch/fairseq/blob/master/examples/roberta/README.glue.md}}.
% Since RoBERTa only reports the ensemble results on the test set, we implemented a RoBERTa with the same pre-training settings (e.g, pre-training steps) of CAPT and made a single-model GLUE leaderboard submission to have a direct comparison.

Table~\ref{tab:glue} presents the performance of representative models on the GLUE benchmark.
We can see that the proposed CAPT obtains best results on most of datasets.
% CAPT outperforms RoBERTa and ELECTRA, which are two state-of-the-art models on most language understanding datasets, by 0.5\% and 0.4\% improvements of average accuracy. 
In more detail, CAPT outperforms RoBERTa that is a very strong baseline model on most language understanding datasets, by 0.6\% improvements of average score. 
Note that CAPT and RoBERTa are nearly identical in terms of the model architecture and fine-tuning hyperparameters, the only difference being the incorporation of contrastive pre-training in CAPT.
Thus we can attribute the improvements of performance on downstream tasks to the role contrastive pre-training plays in learning noise invariant sequence representations.
%except that CAPT incorporates the contrastive pre-training.
%Thus it demonstrates the importance of contrastive pre-training for learning noise invariant sequence representations to improve the performance of the downstream tasks. 
%We believe it is because CAPT 
We believe the reason behind its success is the capability of CAPT to alleviate the pretrain-finetune discrepancy induced by the noise of pre-training. 
In particular, we find that CAPT performs extremely well on natural language inference (RTE, MNLI) which requires a deep understanding of sentence semantics, with 1.0\% absolute improvement of average accuracy over RoBERTa on the dev set. 
% The reason may be that 
This phenomenon can be possibly explained by the fact that our CAPT can better capture the global semantics of the input sequence due to the more effective sentence-level supervision provided by the contrastive training via negative sampling from a memory queue, resulting in superior model performance therein.
More analysis about the influence of memory queue can be found in Section~\ref{sec:ablation_study}.
%We infer that this is because contrastive learning can better capture the global semantics of the input sentence. 
% More deep analysis can be found in Section~\ref{sec: ablation_study}.

\section{Experiments on vision-language tasks}
\label{sec:visual_textual_experiments}
% This section presents the experiments for learning vision-language representations, elaborated on as follows.

\subsection{Implementation}
\label{sec:visual_textual_implementation}

Different from language tasks, the input sequence in the vision-language domain consists of visual region features paired with textual words.
In this scenario, we build CAPT on LXMERT \cite{tan2019lxmert}, a representative cross-modal representation model that separately encodes visual and textual features and then introduces a cross-modal layer to integrate them. 
Same as Section~\ref{sec:textual_implementation}, we construct the corrupted version $\bm{\hat{x}}$ of the original input $\bm x$ by masking part of visual features or textual words.
In addition to the proposed CAPT which can learn sequence-level representations, following \cite{tan2019lxmert}, we also adopt three other pre-training tasks to learn more fine-grained word/region-level representations.
These tasks include: 
masked language modeling (MLM) that predicts the masked words based on the corrupted input $\bm{\hat{x}}$,
masked region modeling (MRM) that predicts the masked visual region objects based on the corrupted input $\bm{\hat{x}}$,
and image-text matching (ITM) that predicts whether the input word sequence is semantically matched with the visual features. 
Due to space limitations, we do not elaborate on the detailed model architecture and these pre-training tasks here. We strongly recommend readers to refer to \cite{tan2019lxmert} for the details.
We also set the queue size to 8192 and the final training loss for visual-linguistic CAPT is defined as the sum of all the above training objectives.

%We build the proposed CAPT on LXMERT~\cite{tan2019lxmert} for learning vision-language representations. 
We use the preprocessed data provided by~\cite{tan2019lxmert}, which mainly includes CoCo~\cite{lin2014microsoftcoco} and Visual Genome~\cite{krishna2017vg}.
%contains 9.18M image-sentence pairs on 180K distinct images. 
%In more detail, 
% The datasets mainly include CoCo~\cite{lin2014microsoftcoco} and Visual Genome~\cite{krishna2017vg}. They also cover train and dev splits from VQA v2.0~\cite{goyal2017vqa2}, GQA balanced version~\cite{hudson2019gqa}, and VG-QA~\cite{zhu2016visual7w}.
%The sentences are split by WordPiece tokenizer and 
Only 36 objects detected by Faster-RCNN are kept for each image.
The model architecture is the same as~\cite{tan2019lxmert}, which consists of 9 language layers, 5 vision layers, and 5 cross-attention layers, with 768 hidden size.
% The temperature $\tau$ is set to 0.8 and $\lambda$ in Eq.~(\ref{eq:lxmert_total}) is 0.5. 
% The pre-trained LXMERT is loaded as a warm-start and trained for another 3 epochs. 
We employ Adam optimizer with the peak learning rate $1e-4$ paired with linear warmup and decay.
%, which is warmed up over the first 5K steps and then linearly decayed with rate 0.1. 
The batch size and dropout are set to 512 and 0.1, respectively. 
%We continue to train the official released LXMERT for the same number of steps as CAPT to have a fair comparison.
%All experiments are conducted on 8 $\times$ 32GB NVIDIA V100 GPUs.

\subsection{Evaluation tasks}
\label{sec:visual_textual_tasks}

%%%%%%%%%%%%%%%%%%%%%%%%%% fig: visual_textual_tasks1 %%%%%%%%%%%%%%%%%%%%%%%%%%%%%%
\begin{table*}[t]
\vskip -0.05in
\footnotesize
\renewcommand{\arraystretch}{1.2}
\begin{center}
\begin{tabular*}{\textwidth}{l@{\extracolsep{\fill}}cccccc}
\toprule
  \multirow{2}{*}{\bf Model} & \multicolumn{2}{c}{\textbf{VQA}} & \multicolumn{2}{c}{\textbf{GQA}} & \multicolumn{2}{c}{\bf $\text{NLVR}^2$}\\ 
  % & \multicolumn{2}{c}{443K/Acc} & \multicolumn{2}{c}{15.4M/Acc} & \multicolumn{2}{c}{86K/Acc} \\
   & test-dev & test-std & test-dev & test-std & dev & test-p \\
\midrule
  %BUTD~\citep{anderson2018bottom}   & 65.32 & 65.67 & -  & -   & -   & -   & -   \\
	SOTA (No pre-training) & 70.63 & 70.90 & 55.8 & 56.1 & 54.80 & 53.50 \\
  ViLBERT~\citep{lu2019vilbert}    & 70.55 & 70.92 & -- & -- & -- & -- \\
  VisualBERT~\citep{li2019visualbert} & 70.80 & 71.00 & -- & -- & 67.40 & 67.00 \\
  VL-BERT~\citep{su2019vlbert}    & 71.79 & 72.22 & -- & -- & -- & --  \\
\midrule
LXMERT~\citep{tan2019lxmert} & 72.42 & 72.54 & 59.95 & 60.33 & 74.82 & 74.41 \\
CAPT (Ours) & \textbf{72.78} & \textbf{73.03} & \textbf{60.48} & \textbf{60.93} & \textbf{75.12} & \textbf{75.13} \\
\bottomrule
\end{tabular*}
\end{center}
\caption{Comparison to the state-of-the-art systems with the single model on VQA, GQA and $\text{NLVR}^2$.
%The information below each task denotes the number of training instances and the evaluation metric. 
The results of both VQA and GQA are reported on the ``test-dev'' split (used for validation on the official server) and the ``test-std'' split (used for maintaining the public leaderboard). The $\text{NLVR}^2$ results are reported on the local dev set (``dev'') and the public test set (``test-p''). The results of baselines except LXMERT are obtained from prior work.}
\label{tab:visual_textual_results}
\vskip -0.05in
\end{table*}
%%%%%%%%%%%%%%%%%%%%%%%%%% fig: visual_textual_tasks1 %%%%%%%%%%%%%%%%%%%%%%%%%%%%%%

We perform evaluation on three benchmark tasks: VQA, GQA, and $\text{NLVR}^2$. VQA~\cite{goyal2017vqa2} aims to select the correct answer based on both the question and its paired image, while GQA~\cite{hudson2019gqa} shares the same task setting but require more reasoning. The goal of $\text{NLVR}^2$~\cite{suhr2019nlvr2} is to predict whether the statement correctly describes the two images. All three tasks use accuracy (Acc) as the evaluation metric.

For VQA and GQA, we add extra multi-layer perceptrons (MLP) that take the representation of \texttt{[CLS]} as the input to perform classification. 
Since each instance in $\text{NLVR}^2$ is composed of two images $(v_1, v_2)$ and a sentence $s$, we use the representation model to encode $(v_1, s)$ and $(v_2, s)$, respectively. Then, a similar MLP takes the concatenation of the \texttt{[CLS]} representations of both $(v_1, s)$ and $(v_2, s)$ as the input to perform classification.
%We apply Adam optimizer to fine-tune our pre-trained model on three tasks for 4 epochs. The batch size on three tasks is 64, 128, and 64, respectively. The corresponding learning rate is set to $5e-5$, $3e-5$, and $5e-5$. 
We adopt the same hyper-parameters with LXMERT for fine-tuning to make a fair comparison.
%The model with the highest accuracy on the dev set are selected as the final model.

\subsection{Results}
\label{sec:visual_textual_results}
%下面是重新写的实验结果
Table~\ref{tab:visual_textual_results} presents the comparison between our approach and several representative systems on the vision-language tasks. We observe the consistent performance boost for our CAPT on all three tasks. 
For instance, it yields a 0.6\% gain over the base architecture LXMERT on GQA and also surpasses various baselines on two other tasks. 
Such improvements indicate the enhanced capability of CAPT to learn noise invariant representations as well as capture the joint semantic representation of the input image-text pair. 
It is also worth noting that the performance gain of our approach on GQA is greater than that on VQA.
The reason may be that GQA pays more attention to visual reasoning, which imposes higher demands on the modeling of joint semantics of visual-textual information. 
Correspondingly, our approach displays competence in the modeling of the global semantics of the input by introducing more effective sentence-level supervision, thereby attaining better results.
Different from VQA and GQA, the goal of $\text{NLVR}^2$ is to determine whether a language caption is true about a pair of images. The increased accuracy on this task demonstrates the universal efficacy of our CAPT under a variety of task settings.

\section{Further analysis}
\label{sec:further_analysis}
% Here we perform a further detailed analysis of our proposed approach, which is elaborately on as follows.

\subsection{Ablation study}
\label{sec:ablation_study}
\begin{figure}
\centering
\includegraphics[width=1.0\textwidth]{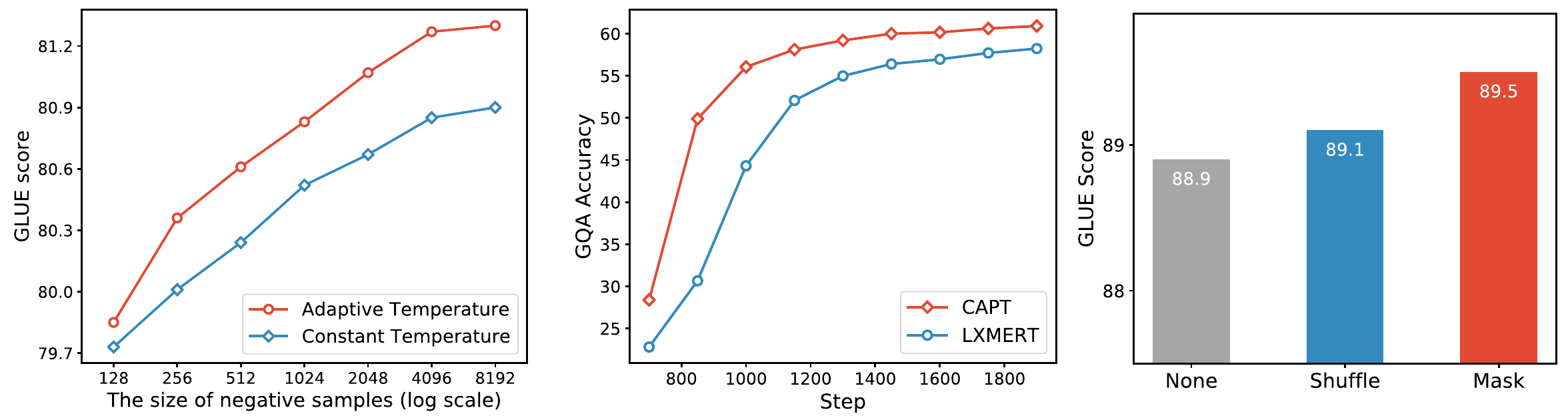}
\vskip -0.05in
\caption{\underline{Left}: The influence of adaptive temperature and the size of negative samples on the GLUE dev set of small CAPT models. \underline{Center}: The validation curves regarding the accuracy of both CAPT and the base architecture LXMERT on the GQA dev set. \underline{Right}: The GLUE dev results of large CAPT models that perform two different ways to construct the corrupted sequence.}
\label{fig:tau+queue}
\vskip -0.05in
\end{figure}

We conduct an ablation study to verify the effectiveness of adaptive temperature and memory queue proposed in Section~\ref{sec:model_extension}.
%, whose results are presented in Figure~\ref{fig:tau+queue}.
As illustrated in Figure~\ref{fig:tau+queue} (Left), fixing the size of negative samples, adaptive temperature exhibits consistent superiority over manually tuned constant temperature. 
Our elaborately designed inverted triangle schedule regarding temperature allows the self-adjustment of the gradients at different stages of contrastive pre-training, leading to a significant gain in model performance.
Figure~\ref{fig:tau+queue} (Left) also demonstrates that the GLUE score of the model consistently increases with the size of stored negative representations, and the absence of memory queue (corresponding to the size of 128 in the figure) could result in a considerable degradation in performance.
As depicted in prior work~\cite{simclr}, large-scale negative samples can assist the model to capture distinctive high-level information of the inputs, therefore enhancing its capacity of feature extraction.

\subsection{Analysis of corruption noise}
\label{sec:noise}

We conduct deeper analysis by constructing $\bm{\hat{x}}$ by means of the shuffling noise to gain further insight into the influence of different corruption methods.
%performed on the corrupted sequence $\bm{\hat{x}}$ in Eq.(\ref{eq:hidden}).
For implementation, we apply a random permutation to the input sequence $\bm x$ within a fixed window $k$ to construct its corrupted version $\bm{\hat{x}}$. Following~\cite{lample18unsupervised,wang2019structbert}, we set the window $k$ to 3. Figure~\ref{fig:tau+queue} (Right) presents the GLUE average score on the dev set when the corrupted sequence $\bm{\hat{x}}$ is constructed by shuffling or masking part of tokens. 
It suggests that both masking and shuffling contribute to the improvement of performance over the baseline which does not perform any corrpution noise to compute contrastive loss. 
The results also reveal the superiority of the masking operation over the shuffling operation.
We speculate that the reason behind this phenomenon may be twofold. 
In the first place, since the baseline model is pre-trained with masked language modeling alone, the learned representations are covariant only with the masking noise. Thus, using the masking noise when applying CAPT to the baseline model makes more sense than using the shuffling noise. 
The other reason may also lie in the observation that masking endows the model with the ability to ``associate'' words, while shuffling only allows the model to learn to ``reorder'' words.
In other words, masking is likely to bring more informative supervision signals than shuffling when learning language representations.
Due to the analysis above, we opt for masking as the noise when implementing our final version of CAPT.

% This may be because the masked perturbation signal allows the model to learn a ``association'' ability, while a shuffled perturbation signal can only allow the model to learn about the ability to ``reorder''.

% Furthermore, performing masking and shuffling at the same time also has a slight improvement on the base model.
% However, the improvements caused by the shuffling supervision on the larger model is covered by masking supervision.

% Meanwhile, considering the input of the masked language modeling is the masked sequence, therefore we simply only perform the masking to conduct corrupted sequence $\bm{\hat{x}}$. In this way, the representations of $\bm{\hat{x}}$ can be used for computing the contrastive loss and masked language modeling loss at the same time, which shares the forward propagation time.

% It demonstrates that shuffling may become less effective for the larger model. The reason may be the improvements caused by the shuffling supervision on the larger model is covered by masking supervision.

%%%%%%%%%%%%
%%%%%%%%%%%%

\subsection{Alleviation of pretrain-finetune discrepancy}
\label{sec:reduce_ptd}

In order to verify that the proposed CAPT can effectively alleviate the pretrain-finetune discrepancy induced by the noise of pre-training, we plot validation curves regarding the accuracy of both CAPT and the base architecture LXMERT on GQA. 
Figure~\ref{fig:tau+queue} (Center) presents the corresponding results, illustrating that the self-supervised model pre-trained by CAPT not only exhibits increased fine-tuning speed but also obtains better performance.
As shown in Figure~\ref{fig:tau+queue} (Center), in the early stage of fine-tuning (the first 1K steps), the proposed CAPT managed to maintain an absolute lead of 5\%-10\% over LXMERT consistently. 
This demonstrates that representations learned by CAPT are more applicable for the data distribution in downstream tasks, rendering model transfer more effective.
By narrowing the differences between representations of the original sequence and its corrupted version, the model is encouraged to learn noise invariant sequence representations. 
In this way, the pre-trained representation model is congruous with the task setting of noiseless inputs in downstream scenarios, leading to better model performance.

\section{Related work}
\label{sec:related_work}

% In summary, this work is mainly related to the following three lines of research.

\paragraph{Pre-trained language representations.}

This task strives to build linguistic representations benefiting various downstream tasks. 
%In terms of model architecture, 
One line of research focuses on autoregressive (AR) pre-training, while the other centers on denoising autoencoding (DAE).
%as the core foundation. 
Representative work of AR pre-training includes ELMo~\cite{peters2018elmo} and GPT~\cite{radford2018improving}, which aim to predict the next word based on previous tokens 
%in a unidirectional pattern 
but lack the modeling of bidirectional context.
Furthermore, XLNet~\cite{yang2019xlnet} remedies this with %generalized AR pre-training based on 
permutation language modeling, but it enlarges the training cost. 
The other research line is built upon DAE, which strives to reconstruct the original sequence based on the corrupted input by jointly attending to both the left and right context.
% The related endeavors share the same model architecture (transformer encoder), with the core difference being pre-training tasks.
Main efforts focus on \textit{token-level} pre-training tasks. For instance, both BERT~\cite{devlin2019bert} and RoBERTa~\cite{liu2019roberta} adopt MLM to recover the masked words.
%or spans~\cite{joshi2019spanbert}. 
StructBERT~\cite{wang2019structbert} 
%attempts to 
incorporates word structure by restoring each shuffled token to its correct position.
% ELECTRA~\cite{clark2020electra} presents a more efficient approach by partially replacing the original input by the sequence from the generator.
However, DAE introduces the noise discarded on downstream tasks during pre-training, which is prone to learn representations covariant with the input noise, leading to the pretrain-finetune discrepancy.
% However, these approaches neglect the modeling of global semantics of the input. 
% Others address this problem by incorporating supervision signals regarding the representation of entire segments through \textit{sentence-level} tasks (e.g. next sentence prediction~\cite{devlin2019bert} or adjacent sentence prediction and ordering~\cite{wang2019structbert, lan2019albert}). 

Besides, most AR and DAE based pre-training tasks neglect the modeling of global semantics of the input. Some DAE based approaches address this problem by incorporating supervisions regarding the entire segment through \textit{sentence-level} tasks (e.g. next or adjacent sentence prediction~\cite{devlin2019bert,wang2019structbert}).
%or adjacent sentence prediction~\cite{wang2019structbert}). 
However, such training relies heavily on the relative position of segments, which suffers from excessively loose semantic connections. Thus, it tends to result in confusing gradient signals.
%In addition, denoising autoencoding is prone to learn representations that are covariant with the input noise of pre-training, 
In comparison, our CAPT encourages the semantic consistency of the original sequence and its corrupted version (e.g. the masked input) via unsupervised contrastive loss. This not only alleviates the pretrain-finetune discrepancy, but also better captures the global semantics of the input.

\paragraph{Pre-trained vision-language representations.}

This direction attempts to build generic representation models for vision-language tasks. 
In terms of model architecture, one research line focuses on \textit{one-stream} BERT-based architecture, which strives to learn generic image-text representations with a unified model.
The corresponding representative work includes VideoBERT~\cite{sun2019videobert}, VisualBERT~\cite{li2019visualbert}, %B2T2~\cite{alberti2019fusion}, 
UNITER~\cite{chen2019uniter}, Unicoder-VL~\cite{li2019unicodervl}, etc.
They usually first combine visual and textual information by projecting them into a common embedding space. The fused representations are then fed into a single transformer to produce the cross-modal features. 
In contrast, the other line such as ViLBERT~\cite{lu2019vilbert} and LXMERT~\cite{tan2019lxmert} focuses on the \textit{two-stream} architecture. 
They first encode visual and textual features by two separate transformers, respectively. 
Then, the co-attentional transformer layers are introduced to allow visual representations to attend to the textual representations and vice versa. 
% Thus they can adapt to the needs of different input processing for each modality and better capture the interactions between multiple modalities.
As for pre-training tasks, different work exhibits commonalities, all focusing on MRM, MLM, and several specific tasks (e.g. ITM).
However, most of these tasks are prone to learning noise covariant representations in the pre-training stage.
Compared with these endeavors, our CAPT benefits the pre-trained model to learn noise invariant vision-language representations via elaborately-designed semantic contrastive loss, thereby bringing better model performance.

\paragraph{Contrastive learning.} 
Contrastive learning is a branch of unsupervised representation learning, which has been widely used in learning graph representations~\cite{bordes13graph, grover16node2vec, velickovic19graph}, word embeddings~\cite{mikolov2013distributed,morin2005hierarchical}, image/video representations~\cite{ wu2018unsupervised,ye2019unsupervised} and structured world models~\cite{kipf2019contrastive}.
The main idea is to construct pairs of related (similar) data as positive samples (e.g., nodes connected by the same edge or images processed by pretext tasks) and pairs of unrelated data as negative samples, and then learn to classify them via the contrastive loss. 
The contrastive loss can come in several forms, including noise contrastive estimation~\cite{Michael2010NCE,oord2018representation,hjelm2019learning}, instance-wise classification~\cite{wu2018unsupervised}, and etc.
It serves as an unsupervised objective to learn feature embeddings where representations of positive samples are concentrated and negative representations are as distant as possible. 
Inspired by these works, we adapt contrastive learning to the natural language and vision-language domains to learn noise invariant sequence representations and demonstrate its effectiveness in improving massive pre-trained models.

\section{Conclusion}
This work presents contrastive pre-training for learning denoised sequence representations in a self-supervised manner.
By enhancing the consistency between representations of the original sequence and the corresponding corrupted version, the pre-trained model is encouraged to learn noise invariant sequence representations. 
On this account, the proposed approach not only alleviates the pretrain-finetune discrepancy induced by the noise of pre-training, but also better captures the global semantics of the input via more effective sentence-level supervision.
Extensive experiments demonstrate the effectiveness and versatility of our approach, which can achieve consistent improvements over baselines in both language and vision-language domains.

\bibliographystyle{plainnat}
\bibliography{neurips_2020.bib}

\end{document}

% \tableofcontents

\begin{appendices}

% \section{Summary of GLUE Tasks}
% Table~\ref{tab:glue_details} shows the details of the each tasks in GLUE benchmark.

% %%%%%%%%%%%%%%%%%%%%%%%%%%%%%%
% \begin{table*}[h]
% \small
% \begin{center}
% % \begin{tabular*}{\textwidth}{l@{\extracolsep{\fill}}ccccccccc}
% \begin{tabular*}{0.96\textwidth}{lccccccccc}
% \toprule
% & \textbf{CoLA} & \textbf{SST} & \textbf{MRPC} & \textbf{STS} & \textbf{QQP} & \textbf{MNLI} & \textbf{QNLI} & \textbf{RTE} \\ 
% \midrule
% \textbf{Task} & SSC & SSC & SPC & SPR & SPC & SPC & SPC & SPC \\
% \textbf{\#Classes} & 2 & 2 & 2 & - & 2 & 3 & 2 & 2 \\
% \textbf{\#Train} & 8.5k & 67k & 3.5k & 5.7k & 363k & 392k & 108k & 2.5k \\
% \textbf{\#Dev} & 1k & 872 & 3.7k & 1.5k & 40k & 20k & 5.7k & 276 \\
% \textbf{\#Test} & 1k & 1.8k & 408 & 1.4k & 391k & 20k & 5.7k & 3k \\
% \textbf{Metrics} & Mcc & Acc & Acc & Pcc & Acc & Acc & Acc & Acc \\
% \midrule 

% \end{tabular*}
% \caption{Summary of the GLUE benchmark. All the tasks are framed as Single Sentence Classification (SSC), Sentence Pair Classification (SPC), and Sentence Pair Regression (SPR).}
% \label{tab:glue_details}
% \end{center}
% \end{table*}
% %%%%%%%%%%%%%%%%%%%%%%%%%%%%%%

% \newpage

\section{Linguistic CAPT Pre-training Details}
Table~\ref{tab:pretraining_hyperparams} describes the hyperparameters for pretraining of the base and large linguistic CAPT models.

% %%%%%%%%%%%%%%%%%%%%%%%%%%%%%%
% \begin{table*}[h]
% \begin{center}
% \begin{tabular}{lcccc}
% \toprule
% \bf Hyperparam  & \bf \textsc{SMALL} & \bf \textsc{BASE} & \bf \textsc{LARGE} \\
% \midrule 
% Number of Layers & 6 & 12 & 24 \\
% Hidden size & 256 & 768 & 1024  \\
% FFN inner hidden size & 1024 & 3072 & 4096 \\
% Attention heads & 4 & 12 & 16 \\
% % Attention head size &  & 64 & 64 \\
% Dropout & 0.1 & 0.1 & 0.1 \\
% % Attention Dropout  & 0.1 & 0.1 & 0.1 \\
% % Warmup Steps & 1k & 1k & 1k \\
% Batch Size & 2k & 2k & 8k \\
% Max Steps & 100k & 10k & 500k \\
% Peak Learning Rate & 5e-4 & 5e-4 & 5e-4 \\
% Learning Rate Decay & Linear & Linear & Linear \\
% Weight Decay & 0.01 & 0.01 & 0.01 \\
% Adam $\epsilon$ & 1e-6 & 1e-6 & 1e-6 \\
% Adam $\beta_1$ & 0.9 & 0.9 & 0.9 \\
% Adam $\beta_2$ & 0.98 & 0.98 & 0.98 \\
% \bottomrule
% \end{tabular}
% \caption{
% Hyperparameters for pretraining base and large Linguistic CAPT models.
% }

%%%%%%%%%%%%%%%%%%%%%%%%%%%%%%
\begin{table*}[h]
\begin{center}
\begin{tabular}{lcccc}
\toprule
\bf Hyperparam  & \bf CAPT-Small & \bf CAPT-Large \\
\midrule 
Number of Layers & 6 & 24 \\
Attention heads & 4 & 16 \\
Hidden size & 256 & 1024  \\
FFN inner hidden size & 1024 & 4096 \\
Aggregation layer inner hidden size & 1024 & 4096 \\
Aggregation layer output hidden size & 256 & 1024 \\
% Attention head size & 64 & 64 \\
Dropout & 0.1 & 0.1 \\
% Attention Dropout  & 0.1 & 0.1 \\
Batch Size & 2k & 8k \\
Max Steps & 200k & 500k \\
Warmup Steps & 10k & 30k \\
Peak Learning Rate & 5e-4 & 6e-4 \\
Learning Rate Decay & Linear & Linear \\
Weight Decay & 0.01 & 0.01 \\
Adam $\epsilon$ & 1e-6 & 1e-6 \\
Adam $\beta_1$ & 0.9 & 0.9 \\
Adam $\beta_2$ & 0.98 & 0.98 \\
\bottomrule
\end{tabular}
\caption{
Hyperparameters for pretraining small and large Linguistic CAPT models.
}
\end{center}
\label{tab:pretraining_hyperparams}
\end{table*}
%%%%%%%%%%%%%%%%%%%%%%%%%%%%%%

% \section{Fine-tuning Details}
% \begin{table*}[t]
% \begin{center}
% \begin{tabular}{lcccccccc}
% \toprule
% \bf Hyperparam  & \textbf{CoLA} & \textbf{SST-2} & \textbf{MRPC} & \textbf{STS-B} & \textbf{QQP} & \textbf{MNLI} & \textbf{QNLI} & \textbf{RTE} \\
% \midrule 
% Learning Rate & 1e-5 & 1.5e-5 & \{1e-5, 2e-5, 3e-5\}\\
% Batch Size & 16 & 48  & \{16, 32\}\\
% Weight Decay & 0.1 & 0.01 & 0.1 \\
% Max Epochs & 4 & 2 & 10 \\
% Learning Rate Decay & Linear &Linear & Linear \\
% Warmup ratio & 0.06 & 0.06 & 0.06 \\
% \bottomrule
% \end{tabular}
% \end{center}
% \caption{
% Hyperparameters for finetuning GLUE.
% }
% \label{tab:roberta_glue_finetune_hyperparams}
% \end{table*}

\end{appendices}

\nobibliography{acl2019}
\bibliographystyle{acl_natbib}